\documentclass{article}
\usepackage{spconf,amsmath,graphicx}
\usepackage{graphicx}
\usepackage{booktabs}
\usepackage{multirow}
\usepackage{color}
\usepackage{cite}
\usepackage{url}
\usepackage{bbm,bm}

\title{Underwater Image Enhancement via Learning Water Type Desensitized Representations}
\name{Zhenqi Fu, Xiaopeng Lin, Wu Wang, Yue Huang, Xinghao Ding $^{\star}$ \thanks{The study is supported partly by the National Natural Science Foundation of China under Grants 82172033, U19B2031, 61971369, 52105126, China Postdoctoral Science Foundation (No.2021M702726), Science and Technology Key Project of Fujian Province(No.2019HZ020009) and Fundamental Research Funds for the Central Universities 20720200003.}}
\address{Key Laboratory of Underwater Acoustic Communication \\ and Marine Information Technology, Ministry of Education, Xiamen University\\
School of Informatics, Xiamen University, China\\
$^{\star}$dxh@xmu.edu.cn}

\begin{document}
%\ninept
%
\maketitle
\begin{abstract}
We present a novel underwater image enhancement method termed SCNet to improve the image quality meanwhile cope with the degradation diversity caused by the water. SCNet is based on normalization schemes across both spatial and channel dimensions with the key idea of learning water type desensitized features. Specifically, we apply whitening to de-correlate activations across spatial dimensions for each instance in a mini-batch. We also eliminate channel-wise correlation by standardizing and re-injecting the first two moments of the activations across channels. The normalization schemes of spatial and channel dimensions are performed at each scale of the U-Net to obtain multi-scale representations. With such water type irrelevant encodings, the decoder can easily reconstruct the clean signal and be unaffected by the distortion types. Experimental results on two real-world underwater image datasets show that our approach can successfully enhance images with diverse water types, and achieves competitive performance in visual quality improvement.
\end{abstract}
\section{Introduction}

Underwater optical vision is a critical perception component for marine research and underwater robotics. For example, underwater surveillance systems and autonomous underwater vehicles rely on high-quality images to fulfill their objectives. Scientists also need clean underwater images to study deteriorating coral reefs and other aquatic life \cite{2}. Unfortunately, the images acquired for these applications are commonly degraded due to various influences. One of the major factors is wavelength-dependent light absorption and scattering over the depth of objects in the scene. The absorption effect is caused by the fact that the red light is absorbed at a higher rate than green and blue in the water. Hence, underwater images are commonly dominated by bluish or greenish tint. The scattering phenomenon (including forward-scattering and backward-scattering) stems from suspending particles, which diminishes the image quality by introducing a homogeneous background noise and haze-like appearance.

\begin{figure}
	\centerline{\includegraphics[scale=0.42]{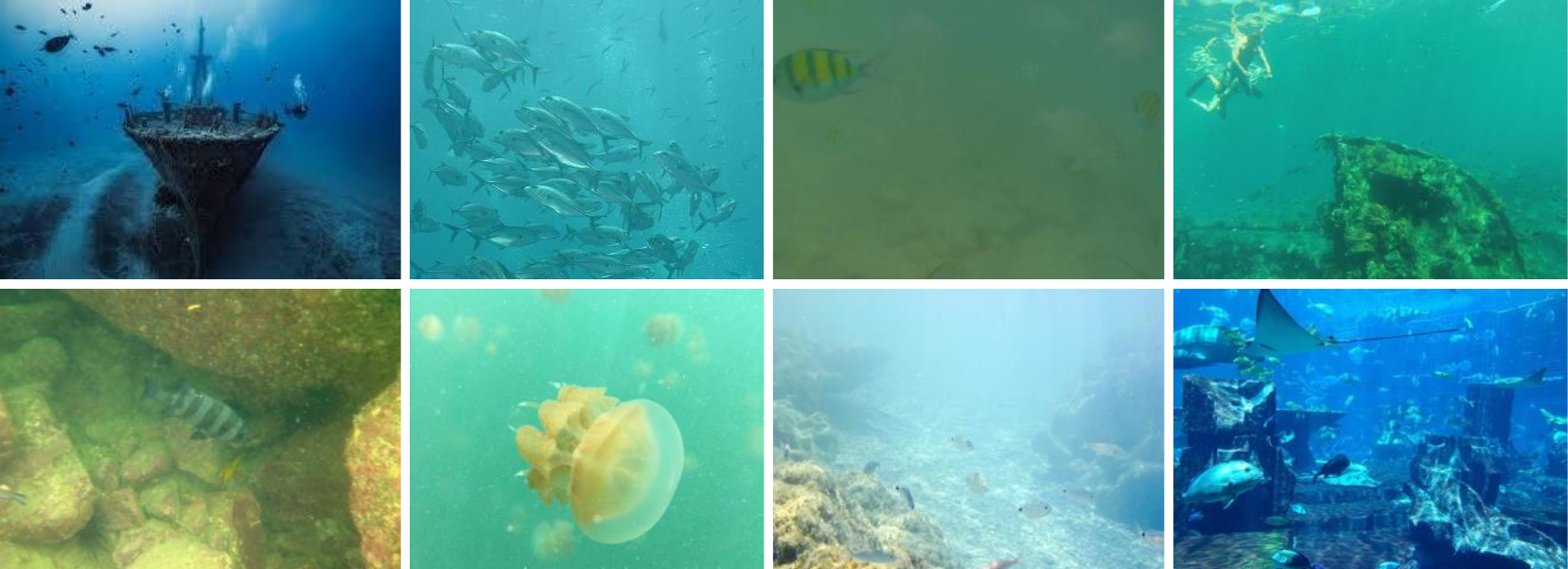}}
	\caption{Underwater scenes captured in diverse water types show a significant difference in appearances and styles.}
\end{figure}

Apart from the light attenuation, another challenge in underwater image enhancement (UIE) is the diversity of degradations. As presented in Fig. 1, underwater scenes captured in diverse water types (e.g., shallow coastal waters, deep oceanic waters, and muddy waters) show a significant difference in appearances and styles. Normally, it is difficult for a single model to enhance underwater images with such multiple degradation distributions. In other words, providing a universal solution for UIE is challenging.

Although numerous UIE approaches have been developed, such as model-free methods \cite{3,4}, prior-based methods \cite{5,6,7,8,23}, and data-driven methods \cite{9,10,11,12}, few of them consider the challenges of degradation diversity explicitly. Anwar et al. \cite{13} first synthesized ten different underwater image datasets, then they trained UIE models for each water type. However, this approach seems inefficient and relies on the prior knowledge of the water type for the given image. Since we do not know the water type ahead of time, Berman et al. \cite{8} tried different parameter sets out of an existing library of water types (e.g., Jerlov water types). Each set leads to a different restored image and the one that best satisfies the Gray-World assumption is chosen as the final output. Similarly, this approach is also inefficient because it must perform ten times for each image to be enhanced. Additionally, it is unreliable to select the best result by a simple Gray-World based quality assessment metric. Recently, Uplavikar et al. \cite{14} learned domain agnostic features for multiple water types and generated clean versions from those features. Concretely, they additionally used a network to classify the water type of a given image from U-Net's \cite{13} encodings. An adversarial loss is used to force the classifier to be unsure of the possible water types. Finally, water type agnostic representations can be learned by the U-Net's encoder.

In this paper, we propose SCNet, a normalization-based approach for UIE to improve the image quality meanwhile handle the diversity of water types. Generally, the degradation diversity of underwater images is introduced by both absorption and scattering. The former makes the image presents different colors. While the latter blurs the edges of objects and leads to various degrees of haze-like effects. Therefore, SCNet normalizes activations across both spatial and channel dimensions, which can reduce the impact of water types. As a result, SCNet can predict a cleaned image more accurately. In summary, this paper introduces the following contributions: 1) We present a normalization-based UIE method. Instead of designing a complex network architecture, we perform normalization schemes at each scale of a simple U-Net to learn multi-scale water type desensitized representations. 2) To obtain better enhancement performance, we not only de-correlate the activations across spatial dimensions but also normalize and re-inject the first two moments of the activation across channel dimensions. 3) Experimental results demonstrate that our approach outperforms the previous methods significantly in both improving the visual quality and dealing with the diversity of water types.

\section{Approach}
Our solution is based on normalization schemes, which standardize and whiten data using the extracted statistics. As shown in Fig. 2, we combine normalization schemes with the U-Net \cite{13} to learn water type desensitized representations. Specifically, we perform spatial-wise and channel-wise normalization at each scale of the U-Net simultaneously. 
\begin{figure*}
	\centerline{\includegraphics[scale=0.275]{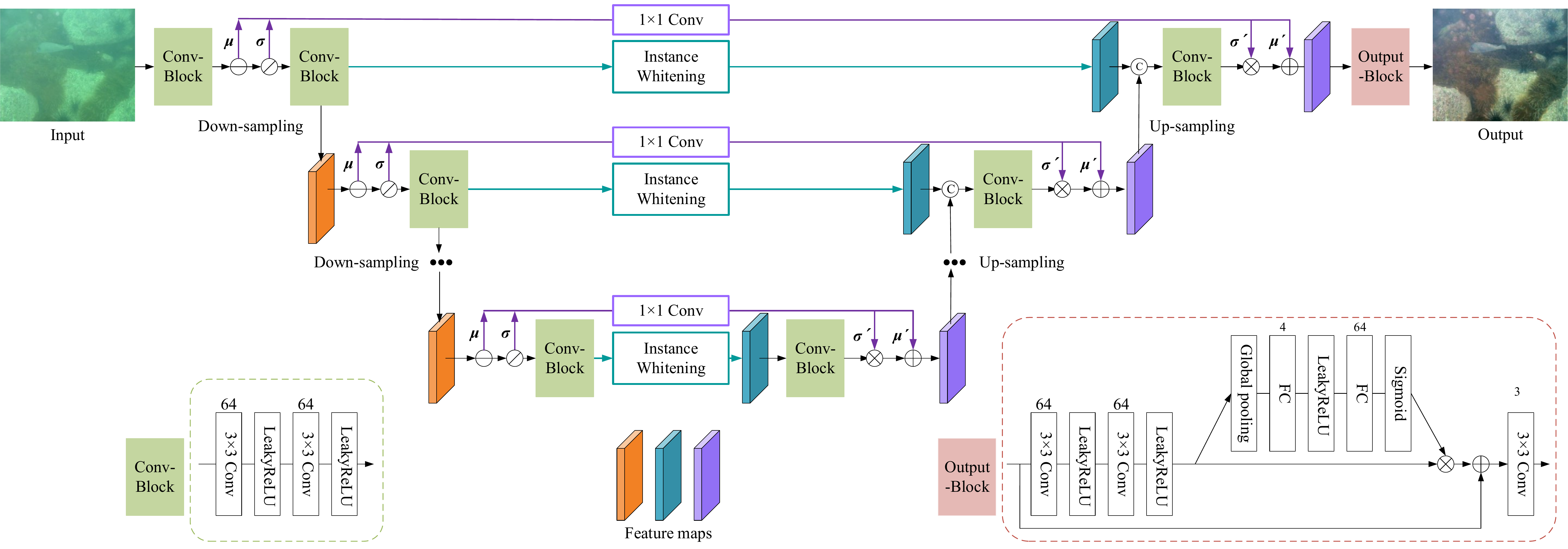}}
	\caption{Illustration of normalization across spatial and channel dimensions at each scale of the U-Net. The normalized activations are style and appearance irrelevant. The decoder can easily reconstruct the clean signal. We embed a SE \cite{24} unit in the output block to improve the network modeling capacity.}
\end{figure*}
\subsection{Spatial-wise Normalization}

Spatial-wise normalization is performed via instance whitening \cite{17} to reduce the influence of diverse water types and discard the extracted statistics across spatial dimensions. We propose to adopt instance whitening to normalize features because the appearance of an individual image can be well encoded by the covariance matrix. In our method, spatial-wise normalization is performed in the U-net's skip-connection. Let ${\bf{X}} \in {\mathbbm{R}^{C \times NHW}}$ refers to the data matrix of a mini-batch, where, $C$, $N$, $H$, $W$ indicate the number of channels, number of instances, the height, and the width respectively. Here, $N$, $H$, and $W$ are viewed as a single dimension for convenience. Let the matrix ${\bf{X}}_n \in {\mathbbm{R}^{C \times HW}}$ be the $n$-th instance in the mini-batch, where $n \in \left\{ {1,2, \ldots ,N} \right\}$. Then the whitening transformation $\Gamma$ for an instance ${\bf{X}}_n$ can be formulated as:
\begin{equation}
	\Gamma \left( {{{\bf{X}}_n}} \right) = {\bm{\Sigma} ^{ - 1/2}}\left( {{{\bf{X}}_n} - \bm{\mu}} \right)
\end{equation}
where $\bm{\mu}$ and $\bm{\Sigma}$ denote the mean vector and the covariance matrix computed from the data, respectively. Specifically, for instance whitening, $\bm{\mu}$ and $\bm{\Sigma}$ are calculated within each individual sample by:
\begin{equation}
	{\bm{\mu}} = \frac{1}{{HW}}{{\bf{X}}_n}
\end{equation}
\begin{equation}
	{\bm{\Sigma}} = \frac{1}{{HW}}\left( {{{\bf{X}}_n} - \bm{\mu}} \right)\left( {{{\bf{X}}_n} - \bm{\mu}} \right)^T + \alpha {\bf{I}}
\end{equation}
where $\alpha$ is a small positive number to prevent a singular ${\bm{\Sigma}}$. In this way, the whitening transformation $\Gamma$ whitens each instance separately (i.e., $\Gamma \left( {{{\bf{X}}_n}} \right)\Gamma {\left( {{{\bf{X}}_n}} \right)^T} = {\bf{I}}$). Note that, in the covariance matrix $\bm{\Sigma}$, the diagonal elements are the variance for each channel, while the off-diagonal elements are the correlation between channels. Therefore, Eq. (1) cannot only standardize but also de-correlate the activations. To enhance the representation capacity, we add scale and shift operations for instance whitening. Thus, Eq. 1 can be rewritten as:
\begin{equation}
	\Gamma \left( {{{\bf{X}}_n}} \right) = {\bm{\Sigma} ^{ - 1/2}}\left( {{{\bf{X}}_n} - \bm{\mu} } \right)\gamma  + \beta
\end{equation}
where $\gamma$ and $\beta$ are learnable parameters denoting the scale and shift operations, respectively.

\subsection{Channel-wise Normalization}
Diverse water types lead to different degrees of scattering effects, which blur the image edge and reduce the visibility of important objects. Considering that channel-wise statistics are position-dependent and can well reveal the structural information about the input image and extracted features \cite{16}, we propose to leverage channel-wise normalization to further reduce the impact of degradation diversity. Concretely, we first remove the mean and the standard deviation across channels in U-Net's encoder. Naturally, the remaining representations are structure irrelevant. Although removing the first two moments does benefit training, it also eliminates important information about the image content, which would have to be painfully relearned in the decoder. Therefore, similar to \cite{16}, we re-inject them into the decoder. Specially, we use $1 \times 1$ convolutional operator to generate optimized statistics (multi-channel outputs). Similar to the notation definition in the spatial-wise normalization, let matrix ${\bf{X}}_n \in {\mathbbm{R}^{HW \times C}}$ be the $n$-th sample in the mini-batch, where $n \in \left\{ {1,2, \ldots ,N} \right\}$. Then the channel-wise normalization $\Omega$ for a sample ${\bf{X}}_n$ can be calculated as:

\begin{equation}
	\Omega \left( {{{\bf{X}}_n}} \right) = \frac{{{{\bf{X}}_n} - \bm{\mu}  }}{\bm{\sigma}}
\end{equation}
where $\bm{\mu}$ and $\bm{\sigma}$ are the mean and standard deviation vectors, respectively. $\bm{\mu}$ and $\bm{\sigma}$ are calculated by:
\begin{equation}
	{\bm{\mu}} = \frac{1}{{C}}{{\bf{X}}_n}
\end{equation}
\begin{equation}
	\bm{\sigma}  = \sqrt {\frac{1}{C}\sum \left( {{{{\bf{X}}_n} - \bm{\mu} } }  \right)^2} + \alpha {\bf{I}}
\end{equation}
where $\alpha$ is a small positive number to prevent a singular ${\bm{\sigma}}$. As mentioned before, after removing the mean and standard deviation across channel dimensions, we transform and re-inject them into the corresponding decoder layer. To be specific, the mean is added to the features, and the standard deviation is multiplied, which can be written as:
\begin{equation}
	{\bf{y}} = \bm{\mu} '{\bf{x}} + \bm{\sigma} '
\end{equation}
where $\bm{\mu} '$ and $\bm{\sigma} '$ denote the optimized statistics of $\bm{\mu}$ and $\bm{\sigma}$, respectively.

\subsection{Loss Function}
Given a training set $\left\{ {{{\bf{u}}_{raw}},{{\bf{u}}_{gt}}} \right\}$, where ${{\bf{u}}_{raw}}$ indicates the raw underwater instances and ${\bf{u}}_{gt}$ refers to the corresponding clean versions. We adopt mean squared error (MSE) and perceptual similarity (PS) \cite{18} for training our network. MSE is calculated based on pixel-wise difference:
\begin{equation}
	{\ell _{{\rm{MSE}}}} = \frac{1}{n}\sum {{{\left( {{{{\bf{\hat u}}}_{gt}} - {{\bf{u}}_{gt}}} \right)}^2}}
\end{equation}
where ${{{\bf{\hat u}}}_{gt}}$ denotes the enhanced output. $n$ is the number of pixels. The perceptual similarity assesses a solution concerning perceptually relevant characteristics. Here, the perceptual similarity is defined as the euclidean distance between the feature representations of enhanced images and clean instances. It can be formulated as follows:
\begin{equation}
	{\ell _{{\rm{PS}}}} = \frac{1}{{{m}}}\sum {{{\left( {\varphi_{i,j} \left( {{\bf{\hat u}}_{gt}} \right) - \varphi_{i,j} \left( {{\bf{u}}_{gt}} \right)} \right)}^2}}
\end{equation}
where ${\varphi_{i,j}}$ indicates the feature map obtained by the $j$-th convolution (after activation) before the $i$-th max-pooling layer within the pr-trained VGG16 network \cite{19}. $m$ is the number of pixels of all feature map extracted. The overall loss function consists of two components and is minimized during the network training. It is expressed as:
\begin{equation}
	{\ell _{{\rm{all}}}} = {\ell _{{\rm{MSE}}}} + \lambda {\ell _{{\rm{PS}}}}
\end{equation}
where $\lambda = 0.1$ denotes the weight.

\begin{figure*}
	\centerline{\includegraphics[scale=0.208]{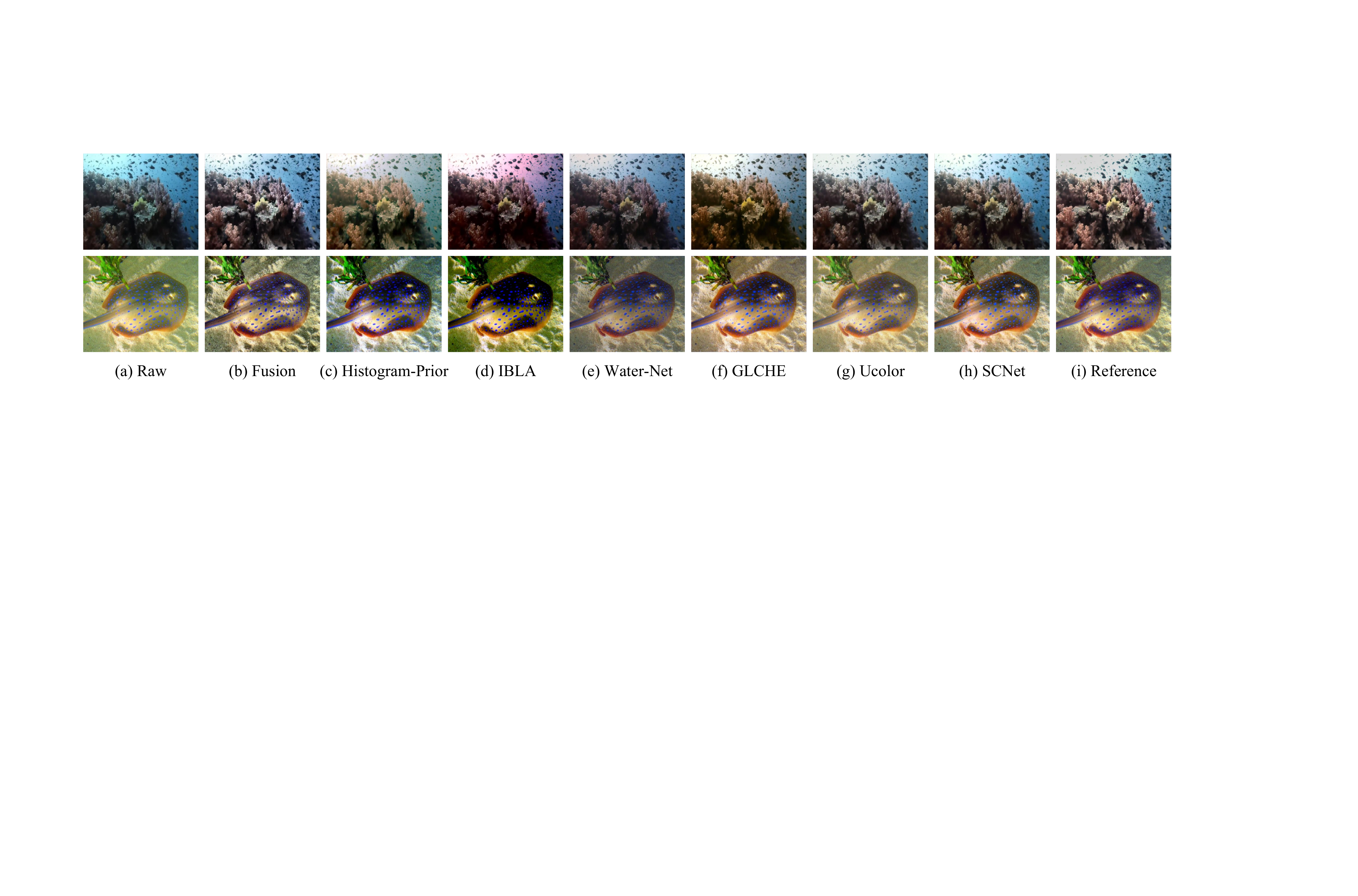}}
	\caption{Visual comparisons on the UIEBD dataset. }
\end{figure*}

\begin{figure}[h!]
	\centerline{\includegraphics[scale=0.451]{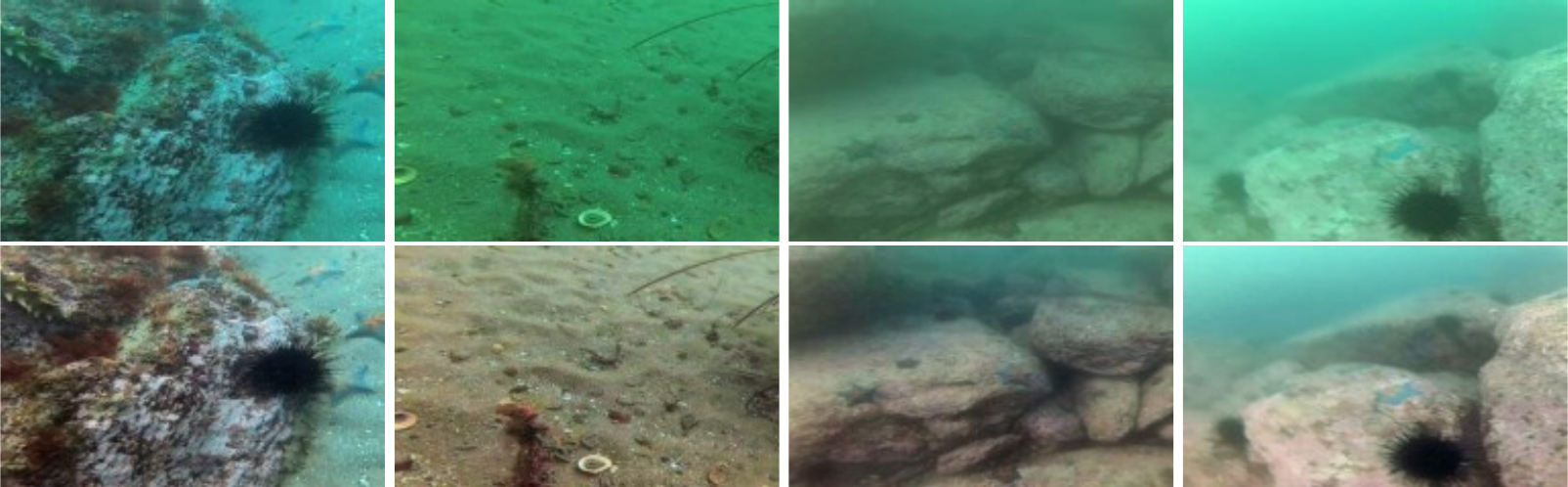}}
	\caption{Visual results on the RUIE dataset.}
\end{figure}

\section{Experiment}

We employ the real-world UIE dataset (UIEBD)\cite{10} to train and test our model. UIEBD contains 890 underwater images and corresponding labels. Note that the reference image in UIEBD is obtained by subjective selections among 12 UIE results. We use the first 700 images for training and the rest for testing. We adopt the PyTorch framework to train our network using Adam solver with an initial learning rate of 1e-4. The mini-batch size is set as 1 empirically. The patch size is $128 \times 128$. The scale of the U-Net is set as 4 (i.e., we down-sample the features 3 times). We compare SCNet with six state-of-the-art UIE methods including one model-free approaches (Fusion \cite{4}), two prior-based approaches (Histogram-Prior \cite{7} and IBLA \cite{22}), and three data-driven approaches (Water-Net \cite{10}, GLCHE \cite{11}, and Ucolor \cite{12}). We use three objective quality assessment metrics (i.e., SSIM, PSNR, and LPIPS \cite{21}) as the performance criteria. A better UIE approach should have higher SSIM and PSNR scores, and a lower LPIPS score. Besides, we test the model performance on another real-world dataset (RUIE) \cite{2} to further demonstrate the model generalization performance. Note that RUIE does not contain reference images. Therefore, only qualitative results are presented on this dataset.

\subsection{Performance Comparisons}

\begin{table}[!t]
	\caption{Performance comparisons on the UIEBD dataset.}\label{tbl1}
	\renewcommand\tabcolsep{9pt}
	\centering	
	\begin{tabular}{cccc}
		\toprule
		Method &  SSIM $\uparrow$  & PSNR $\uparrow$  & LPIPS $\downarrow$ \\ 
		\midrule	
		Fusion   & 0.8222 & 21.1849 & 0.2083\\	 
		Histogram-Prior   & 0.7620 & 18.5148 & 0.3360\\
		IBLA   & 0.5733 & 14.3856 & 0.4299\\		
		Water-Net  & 0.8303 & 19.3134 & 0.2016\\
		GLCHE   & 0.8487 & 21.0270 & 0.1993\\	
		Ucolor	& 0.8395 & 21.6463 & 0.1970\\	
		\textbf{SCNet}  & \textbf{0.8625} & \textbf{22.0816} &  \textbf{0.1936} \\							
		\bottomrule
		
	\end{tabular}
\end{table}

\begin{table}[!t]
	\caption{Ablation study on the UIEBD dataset.}\label{tbl2}
	\renewcommand\tabcolsep{9pt}
	\centering	
	\begin{tabular}{cccc}
		\toprule
		Method &  SSIM $\uparrow$ & PSNR $\uparrow$ & LPIPS $\downarrow$\\ 
		\midrule	
		U-Net   & 0.8350 & 19.5114 & 0.2640 \\
		SCNet w/o SN   & 0.8436 & 20.9934 & 0.2155 \\	 
		SCNet w/o CN  & 0.8579 & 21.3433 & 0.2070\\
		
		\textbf{SCNet (FULL)}   & \textbf{0.8625} & \textbf{22.0816} &  \textbf{0.1936} \\							
		\bottomrule
		
	\end{tabular}
\end{table}

Quantitative results of different UIE algorithms on the UIEBD dataset are presented in Table 1. As we can observe, SCNet achieves the best performance in terms of three full-reference image quality evaluation metrics. Prior-based methods obtain low SSIM and PSNR scores. The reason may line in that prior-based methods rely on handcrafted imaging models and prior features. Model-free methods and data-driven methods achieve higher performance compared with prior-based methods, but all of them are inferior to our method since they ignore the impact of diverse water types. We also provide subjective comparisons in Fig. 3. From the visual results, we observe that SCNet can handle the diversity of water types, and can consistently generate natural and vivid results on testing images. Fig. 4 reports the visual results on the RUIE dataset. As can be observed, SCNet enables generating enhanced images with natural colors and contrasts. This demonstrates that the proposed method has good generalization performance for real-world applications.

\subsection{Ablation Study}
We conduct ablation studies to verify the effectiveness of the proposed spatial-wise and channel-wise normalization schemes. Table 2 presents the test results using four different settings. Due to the limited space, we use “SN” and “CN” to represent “Spatial-wise Normalization” and “Channel-wise Normalization,” respectively. We can observe that directly using U-Net cannot obtain satisfactory results because it does not take the special distortions of the underwater environment into account. Normalizing representations on either spatial or channel dimensions can significantly improve enhancement performance. As expected, the best results are obtained by simultaneously normalizing features on both spatial and channel dimensions. This is because the diversity of water types exists in both spatial and channel dimensions.

\section{Conclusion}

We propose a novel data-driven method for underwater image enhancement. Different from most existing approaches that focus on designing complex network architectures, we propose to combine a simple U-Net with spatial-wise and channel-wise normalization to deal with the diversity of water types. By normalizing activations across both spatial and channel dimensions, appearance irrelevant representations can be effectively learned. As a result, the network can easily reconstruct the clean signal from those latent representations. Experimental results show that SCNet achieves competitive performance and has better generalization ability.

\vfill\pagebreak

\bibliographystyle{IEEE}
\bibliography{cas-refs}

\end{document}